\documentclass[sigconf]{acmart}

\AtBeginDocument{%
  \providecommand\BibTeX{{%
    \normalfont B\kern-0.5em{\scshape i\kern-0.25em b}\kern-0.8em\TeX}}}

\usepackage{epsfig}
\usepackage{graphicx}
\usepackage{amsmath}
\usepackage{multirow} 
\usepackage{booktabs} 
\usepackage{subfigure}
\usepackage{float}
\usepackage{caption}
\usepackage{multicol}
\usepackage{hyperref}
\usepackage{balance}
\hypersetup{colorlinks,linkcolor=red,anchorcolor=blue,citecolor=green,urlcolor=blue}

\newcommand{\minisection}[1]{\vspace{0.04in} \noindent {\bf #1}\ \ }

\usepackage{fancyhdr}
\iftrue

\newcommand{\jing}[1]{\textcolor{red}{\bf [MD: #1]}}
\newcommand{\yu}[1]{\textcolor{blue}{\bf [AG: #1]}}
\newcommand{\LZ}[1]{\textcolor{cyan}{\bf [LZ: #1]}}

\else
\newcommand{\jing}[1]{}
\newcommand{\yu}[1]{}
\newcommand{\LZ}[1]{}
\newcommand{\alertJW}[1]{}
\newcommand{\alertJW}[1]{}
\fi

\copyrightyear{2021}
\acmYear{2021}
\setcopyright{acmcopyright}
\acmConference[MM '21]{Proceedings of the 29th ACM International Conference on Multimedia}{October 20--24, 2021}{Virtual Event, China}
\acmBooktitle{Proceedings of the 29th ACM International Conference on Multimedia (MM '21), October 20--24, 2021, Virtual Event, China}
\acmPrice{15.00}
\acmDOI{10.1145/3474085.3475387}
\acmISBN{978-1-4503-8651-7/21/10}

\begin{document}
\fancyhead{}

\title{Unsupervised Cross-Modal Distillation for Thermal Infrared Tracking}

\author{Jingxian Sun}
\authornote{Both authors contributed equally to this research.}
\authornote{National Engineering Laboratory for Integrated Aero-Space-Ground-Ocean Big Data Application Technology (ASGO).}
\affiliation{
  \institution{ASGO, School of Computer Science, Northwestern Polytechnical University}
  \city{Xi'an}
  \country{China}}
\email{jingxiansun@mail.nwpu.edu.cn}

\author{Lichao Zhang}
\authornotemark[1]
\affiliation{
  \institution{Aeronautics Engineering College, Air Force Engineering University}
  \city{Xi'an}
  \country{China}}
\email{lichao.zhang@outlook.com}

\author{Yufei Zha}
\authornote{Corresponding author.}
\authornotemark[2]
\affiliation{%
  \institution{ASGO, School of Computer Science, Northwestern Polytechnical University}
  \city{Xi'an}
  \country{China}}
\email{yufeizha@nwpu.edu.cn}

\author{Abel Gonzalez-Garcia}
\affiliation{%
  \institution{wrnch}
  \city{Montreal}
  \country{Canada}}
\email{abel.gonzalezgarcia@wrnch.ai}

\author{Peng Zhang}
\authornotemark[2]
\affiliation{%
  \institution{ASGO, School of Computer Science, Northwestern Polytechnical University}
  \city{Xi'an}
  \country{China}}
\email{zh0036ng@nwpu.edu.cn}

\author{Wei Huang}
\affiliation{%
  \institution{School of Information Engineering, Nanchang University}
  \city{Nan Chang}
  \country{China}}
\email{huangwei@ncu.edu.cn}

\author{Yanning Zhang}
\authornotemark[2]
\affiliation{%
  \institution{ASGO, School of Computer Science, Northwestern Polytechnical University}
  \city{Xi'an}
  \country{China}}
\email{ynzhang@nwpu.edu.cn}

\renewcommand{\shortauthors}{Sun, Zhang and Zha, et al.}

\begin{abstract}	
The target representation learned by convolutional neural networks plays an important role in Thermal Infrared (TIR) tracking.
Currently, most of the top-performing TIR trackers are still employing representations
learned by the model trained on the RGB data. However, this representation does not take into account the information in the TIR modality itself, limiting the performance of TIR tracking.	

To solve this problem, we propose to distill representations of the TIR modality from the RGB modality with  Cross-Modal Distillation (CMD) on a large amount of unlabeled paired RGB-TIR data. 
We take advantage of the two-branch architecture of the baseline tracker, \emph{i.e.} DiMP, for cross-modal distillation working on two components of the tracker. 
Specifically, we use one branch as a teacher module to distill the representation learned by the model into the other branch. 
Benefiting from the powerful model in the RGB modality, the cross-modal distillation can learn the TIR-specific representation for promoting TIR tracking.
The proposed approach can be incorporated into different baseline trackers conveniently as a generic and independent component.
Furthermore, the semantic coherence of paired RGB and TIR images is utilized as a supervised signal in the distillation loss for cross-modal knowledge transfer. 
In practice, three different approaches are explored to generate paired RGB-TIR patches with the same semantics for training in an unsupervised way. 
It is easy to extend to an even larger scale of unlabeled training data.	
Extensive experiments on the LSOTB-TIR dataset and PTB-TIR dataset demonstrate that our proposed cross-modal distillation method effectively learns TIR-specific target representations transferred from the RGB modality. Our tracker outperforms the baseline tracker by achieving absolute gains of \textbf{2.3\%} Success, \textbf{2.7\%} Precision, and \textbf{2.5\%} Normalized Precision respectively. Code and models are available at \url{https://github.com/zhanglichao/cmdTIRtracking}.
\end{abstract}



\maketitle
\vspace{-1mm}
\section{Introduction}
\begin{figure}[ht]
	\centering
	\vspace{-3mm}
	\includegraphics[width=0.45\textwidth]{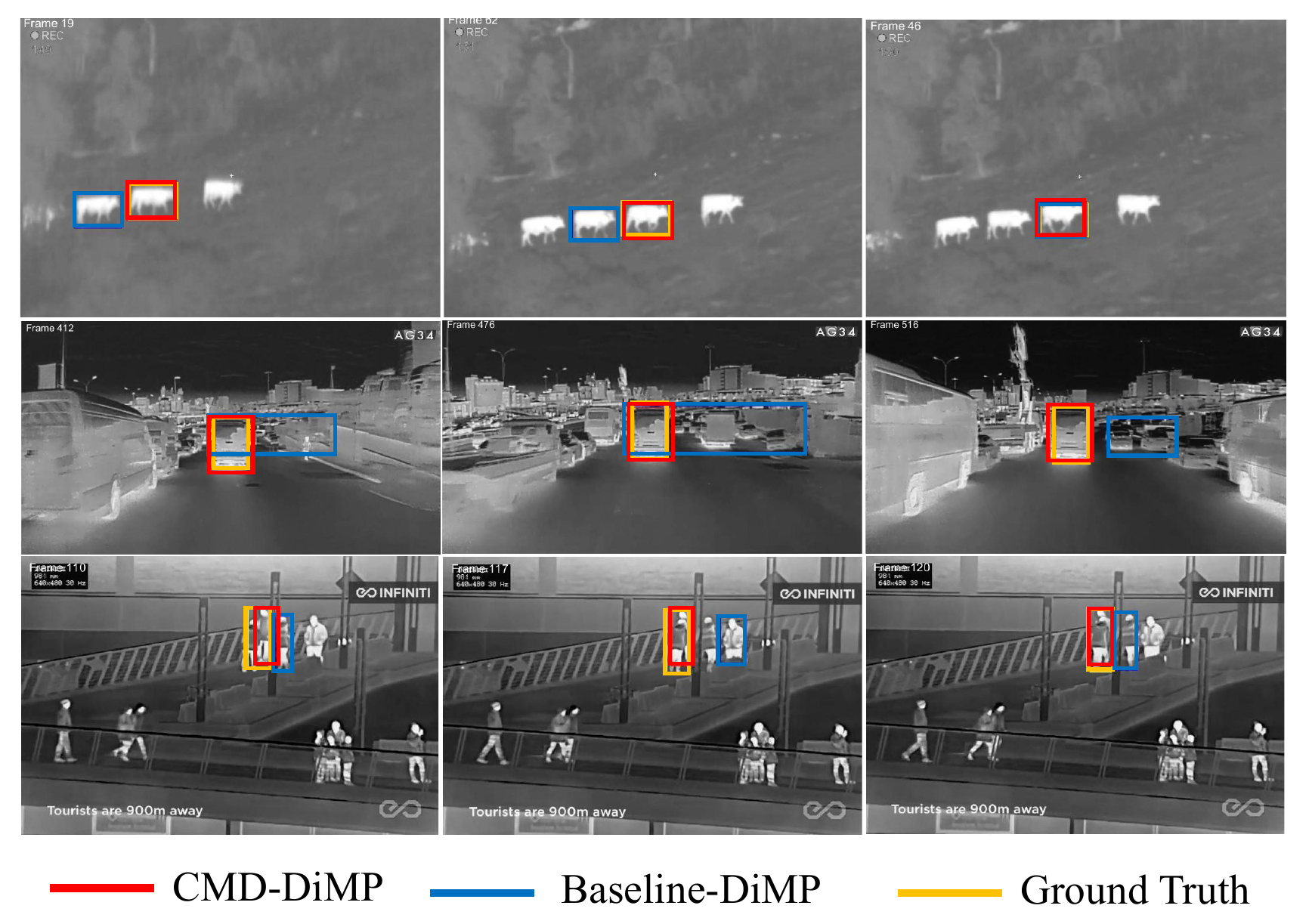}\\
	\vspace{-3mm}
	\caption{Qualitative comparison between the proposed method and baseline tracker DiMP~\cite{DIMP} on LSOTB-TIR dataset~\cite{LSOTB}. 
				By training with cross-modal distillation, the tracker can effectively track the targets on different challenging scenes.
	}
	\vspace{-4mm}
	\label{fig:results}
\end{figure}

Thermal Infrared (TIR) {t}racking~\cite{DSS-tir,SRDCFir,MCFTS2017,MMNet2020} aims to locate a target by using videos of the thermal infrared modality, where the initial {location} of the target {is} given in the first frame. Compared with RGB trackers,  
TIR trackers can still distinguish the target from the background under some challenging situations, ~\emph{e.g.} low illumination, shadow, and occlusion, even working well in total darkness where visual cameras have no signal. These advantages make TIR trackers running in a wide range of applications, such as video surveillance, maritime rescue, various defense systems, and driver assistance at night~\cite{RGBT210}. 
However, one key issue of TIR trackers is how to {learn} powerful features for representing the target efficiently {in order} to deal with various difficulties {specific to the TIR modality}, such as thermal crossover, intensity variation, and distractor{s}.

{In early work}, the hand-craft{ed} features~\cite{DSLT, SRDCFir}, \emph{e.g.} Histogram of Oriented Gradient{s} (HOG)~\cite{HOG}, Harr-like feature~\cite{Sparse-tir}, are employed to represent targets in TIR trackers. These trackers exploit classic learning paradigms, such as multiple instance 
learning~\cite{inproceedings}, discriminative correlation filter~\cite{ASHA2017114}, and low-rank sparse learning~\cite{DBLP:journals/lgrs/HeLZY16} for TIR tracking. 
Recently, Convolution{al} Neural Network{s} (CNN) {have also} been introduced to represent the target to improve the performance of TIR trackers~\cite{MLSSNet2020,MMNet2020}.
The CNN features extracted by the pre-trained networks, \emph{e.g.} VGGNet~\cite{vgg14}, ResNet~\cite{DBLP:conf/cvpr/HeZRS16}, are integrated into the trackers followed by the correlation filter (CF)~\cite{KCF}, structural support vector machine~\cite{LMSCO2018} {or} Siamese networks~\cite{SiamFC} for {enhancing} the target representation.
The results show that the representation with CNN features is discriminative to distinguish the target from {distractors} in the background, compared with hand-crafted features.

However, both hand-crafted and off-the-shelf features are not optimal for the TIR target, which limits the performance of TIR tracking.
In practice, TIR images lack color information and rich texture features compared with RGB images.
The hand-crafted features are designed according to the characteristics of RGB images, while the pre-trained models are derived from large-scale RGB data. 
They have exclusively used the RGB modality, ignoring the differences between RGB and TIR modalities.
As a result, this {gap in the} representation degrades the discriminability of the tracker for identifying the target from the background {in the TIR modality}.  
A TIR-specific representation needs to be specifically tailored for the TIR modality in order to maximally leverage its characteristics. 
Additionally, {the lack of} large-scale annotated TIR data {makes it impossible} to train the networks from scratch. 
Labeling the TIR data is a time-consuming and laborious work. And at present there is no large-scale TIR data for tracking as normally RGB trackers do~\cite{SiamFC,Lasot,Got-10k,ATOM}. Therefore, in our previous work~\cite{ECO-stir}, we collect a large amount of TIR data from other vision tasks, but these TIR data is not annotated for the tracking task. To solve this issue, we propose to use image-to-image translation models to generate synthetic labeled TIR datasets transferring from RGB tracking datasets. Exhaustedly, a lot of extra efforts are still needed for training translation models in this work.

In order to obtain powerful target representations for TIR  tracking,  we propose to distill the representation of the TIR modality from the RGB modality with  Cross-Modal Distillation (CMD) on a large amount of unlabeled paired visible and infrared images. Motivated by the idea of distilling the network knowledge from {a} teacher model to {a} student model~\cite{KD}, our method does distill representation knowledge from the RGB modality to the TIR modality.
We use DiMP~\cite{DIMP} as our baseline tracker which is constructed with the architecture of two branches. We explore distillation operations on Target Center Location (TCL) and Bounding Box Estimation (BBE) in DiMP~\cite{DIMP}. 
As a result, we can obtain the TIR-specific representation guided by the pre-trained {model} in the RGB modality.
Benefiting from the powerful model trained on large-scale labeled RGB data, the learned TIR model can better represent the TIR target.
Here, the proposed method is generic and can be applied to different baseline trackers conveniently. In this work, we use our cross-modal distillation method to train the tracker ATOM~\cite{ATOM} in section~\ref{4_exp_lsotb}, and it effectively improves the performance of ATOM~\cite{ATOM}

Moreover,  an unsupervised training method is proposed to take advantage of the dual-modalities data without any annotations, and this will relieve the dependency on the labeled TIR data.
The semantic coherence of the paired RGB and TIR image replaces the manual labels as the ground-truth in the final loss function for model training.  
This prior is helpful to transfer the model knowledge between the different modalities.
In practice, we explore three different approaches (`center area', `random sampling', and `detection') to generate paired RGB-TIR patches from the RGB and TIR images as training data.
These paired dual-modalities image patches with the same semantics are fed into the network to learn TIR representation under the distillation loss.
Here, we do not require any kinds of annotations in the training data, so the whole training procedure can be conducted in an unsupervised manner. 
Besides, it is easy to extend to an even larger scale of unlabeled training data.

We validate the proposed method on two standard test datasets: LSOTB-TIR dataset~\cite{LSOTB} and PTB-TIR dataset~\cite{PTB-TIR}.
Some qualitative results for comparisons between the proposed method and baseline tracker are shown in Fig.~\ref{fig:results}. 
Compared with the baseline tracker, we achieve absolute gains of 2.3\% Success, 2.7\% Precision and 2.5\% {Normalized Precision} 
respectively on LSOTB-TIR dataset~\cite{LSOTB}.
These results demonstrate that our proposed Cross-Modal Distillation (CMD) method effectively learns TIR-specific target representations transferred from the RGB modality. 
The contributions of our work are as follows:
\vspace{-1mm}
\begin{itemize}
	\item
	{
		A representation transferring approach called Cross-Modal Distillation (CMD) is proposed to distill {a} TIR-specific representation from the RGB modality on a large amount of unlabeled paired RGB-TIR data. 
		This benefits from the powerful model trained on large-scale labeled RGB data. 
		The proposed approach can be incorporated into different baseline trackers conveniently due to its generality and independence.
	}
	\item 
	{
		An unsupervised manner is proposed without any {annotation} of the target for training.
		During training, three different approaches are explored to generate paired RGB-TIR patches with the same semantics automatically.		
		It is easy to extend to an even larger scale of unlabeled training data.
	} 	
	\item {We conduct extensive experiments on two benchmarks to verify the effectiveness of the proposed method.
		The results demonstrate that our algorithm achieves a significant improvement against SOTA methods on the TIR tracking challenge.}
\end{itemize}

%

\vspace{-1.5mm}
The remainder of the paper is structured as follows. 
In section~\ref{sec:related}, we briefly discuss related works. 
In section~\ref{sec:method}, we describe the cross-modal distillation modules to learn the TIR-specific representation transferred from the pre-trained model on the RGB data. 
In section~\ref{sec:exp}, extensive experiments are carried out on two standard thermal infrared tracking datasets. 
Finally, we conclude our work and propose future research plans in section ~\ref{sec:con}.

\vspace{-1.5mm}
\section{Related Work}
\label{sec:related}
In this section, we will introduce the works closely related to our study in this paper. 
More references about multi-modal tracking can be seen in the surveys ~\cite{Multi-modal, comprehensive}.
\vspace{-1mm}
 \begin{figure*}[!htbp]
 	\centering
 	\includegraphics[width=0.8\textwidth]{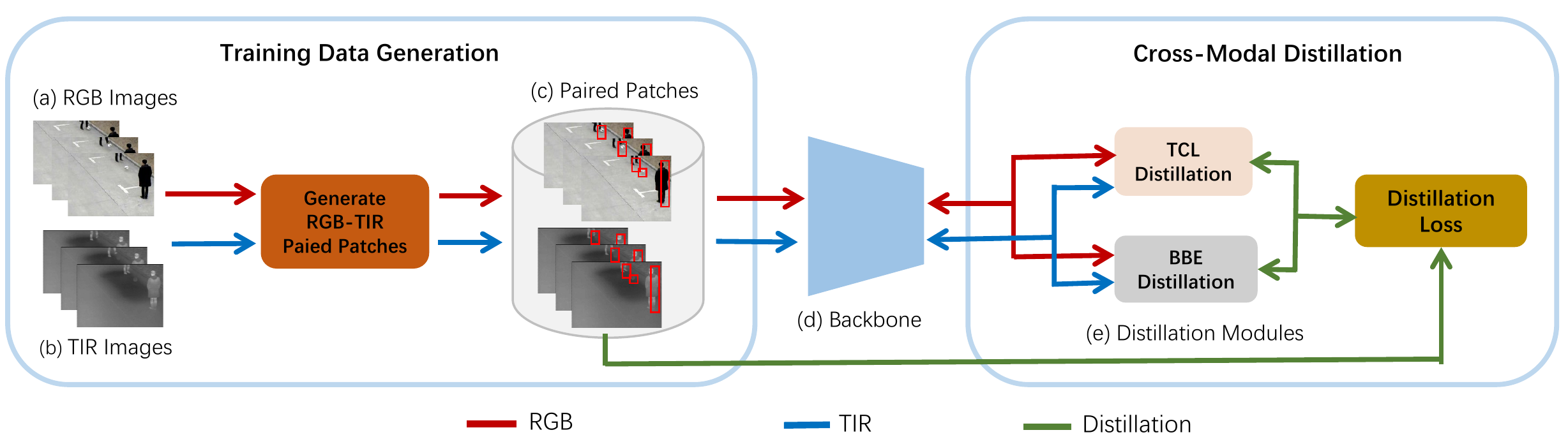}
 	\vspace{-3mm}
 	\caption{Unsupervised training of Cross-Modal Distillation (CMD) for TIR tracking.
 		Only the paired RGB images shown in Fig. \ref{fig:pipeline} (a)  and TIR images shown in Fig. \ref{fig:pipeline} (b)  are employed as training data.
 		Before network training, the paired dual-modalities patches in Fig. \ref{fig:pipeline} (c) are generated automatically as the input of the network for cross-modal distillation. 
 		The CNN features are extracted by backbone network on both RGB and TIR modalities in Fig. \ref{fig:pipeline} (d), and then they are fed into the cross-modal distillation shown in Fig. \ref{fig:pipeline} (e) with target center location (TCL) distillation and bounding box estimation (BBE) distillation. 
 		Here, the red line and blue line denote the processing flow of RGB modality and TIR modality, respectively. The cross-modal distillation flow is expressed as the green line. 
 		The line with the reverse arrow indicates the training process.
 	}
 	\label{fig:pipeline}
 	\vspace{-5mm}
 \end{figure*}

\minisection {TIR Tracking.} The hand-crafted features, such as edges, motion features, and HOG, were integrated discriminative correlation filter with scale estimation~\cite{DSS-tir} or spatial regularization~\cite{SRDCFir} for TIR tracking.
Their favorable performance was mainly due to the robust feature representation and online learning. 
The CNN features extracted from the common networks were used to replace the hand-crafted features for target representation in
TIR tracking~\cite{MCFTS2017}.
Recently,  MLSSNet~\cite{MLSSNet2020} trained a multi-level similarity-based Siamese network on an RGB and TIR dataset simultaneously.
A multi-task matching framework~\cite{MMNet2020} was proposed to learn deep features in the levels of inter-class and intra-class respectively. 
These kinds of
features complemented each other and recognized TIR objects
in the levels of inter-class and intra-class respectively. These
feature models were learned and jointly optimized on the TIR tracking task.
Besides, the Siamese networks were used to extract the features from the network trained on a large amount of synthetic TIR images for TIR tracking~\cite{ECO-stir}.

Unlike previous works,
we expect to learn the TIR-specific discriminative representation transferred from the RGB modality by training the network on the paired RGB and TIR images.

\vspace{-1mm}
\minisection{Knowlege Distillation for Vision Tasks.}
The main idea of knowledge distillation~\cite{KD} was that the student model mimics the teacher model in order to obtain a competitive or even a superior performance, which benefits the deployment of deep neural networks in mobile devices and embedded systems.
In fact, it was important to transfer knowledge between different modalities, because the data or labels of some modalities might not be available during training. 
The idea is to transfer the annotation or label data through pair-wise samples and this has been widely used for cross-modal applicaitons~\cite{CRD}. 
Tian \emph{et al.}~\cite{CRD}  proposed a contrastive loss to transfer pair-wise relationship across different modalities, while GANs were employed to perform cross-modal distillation among the missing and available modalities~\cite{RohedaRKD18}.
In addition, in the field of visual question answering,  the knowledge from trilinear interaction teacher model with image-question-answer as inputs was distilled into the learning of a bilinear interaction student model with image-question as inputs~\cite{DoTDTT19}.
Besides, lots of cross-modal distillation methods~\cite{KunduLR19, ChenL0H19} also transferred the knowledge among multiple domains.

Unlike the above applications, we focused on TIR tracking, which lacked large-scale annotated TIR data for training. 
To overcome this problem, cross-modal distillation (CMD) was introduced to transfer representations from the RGB  modality to the TIR modality in this study. 
Specifically,  the knowledge of the RGB model was transferred to the TIR model through unsupervised learning, and it benefited to improve the performance of the TIR tracking task.

\vspace{-2mm}
\section{Proposed Method}
\label{sec:method}

In this section, we propose to transfer the representation of the pre-trained model from RGB modality to TIR modality by using a large amount of paired RGB-TIR data in an unsupervised way. 
Unlike the classic KD work in~\cite{KD} which uses two independent models for distillation happening on intermediate features or final outputs, our proposed cross-modal distillation adopts representations of the RGB modality as powerful supervision signals in one branch to guide the representation learning in the TIR modality with the other branch.

\vspace{-3mm}
\subsection{Overview}
The training pipeline of the proposed unsupervised training of cross-modal distillation for TIR tracking is shown in Figure~\ref{fig:pipeline}.
In order to transfer the representation from the  RGB modality to the TIR modality, cross-modal distillation modules are discussed in Fig.~\ref{fig:pipeline} (e) following after the backbone network shown in  Fig.~ \ref{fig:pipeline} (d).

Here, the proposed cross-modal distillation modules are constructed by the convolutional layer, pooling layer, fully connected layer, distillation operation, and so on and contains two components for distillation:  the Target Center Location (TCL) distillation for discriminating the target from the background and Bounding Box Estimation (BBE) distillation for fine-tuning the bounding box of the target.
Both these two distillations are guided by representations of the RGB modality from one branch to learn the TIR-specific representations with the other branch under the distillation loss.
Unlike the training procedure of RGB trackers, our approach is trained to process the information from both RGB and TIR modalities simultaneously, which are denoted as red and blue lines respectively in Fig.~ \ref{fig:pipeline}. While the green line represents the procedure of the cross-modal distillation in Fig.~ \ref{fig:pipeline}.
The input of the proposed model architecture are paired dual-modalities images. 
Before the training of the network, three different approaches are explored to generate paired patches shown in Fig. ~\ref{fig:pipeline} (c) and they are from the RGB images as in Fig. ~\ref{fig:pipeline} (a) and TIR images as in Fig. ~\ref{fig:pipeline} (b).
After that, patches' coordinates in the image and pseudo Gaussian distribution are utilized to construct the annotated label for calculating the loss during the training of the cross-modal distillation.

Mathematically, we denote the RGB image as $I^v$ and TIR image as $I^t$, respectively. 
The bounding box is described as $s = (c_x/w, c_y/h, \log $ $w, \log h) \in \mathbb R^4$, where $(c_x,c_y)$ is image coordinates of the bounding box center, the width and height of box are denoted as $w$ and $h$. 
The CNN features extracted by the backbone network are written as $x^v \in \chi $ for RGB modality and $x^t \in \chi$ for TIR modality. 
The training dataset can be denoted as $S_{train} = \{ x_i^v, x_i^t \}_{i=1}^N$, where $N$ is the volume of the training dataset. 

Our goal is to learn a cross-modal distillation $\mathbb{D}(\phi,\theta)$ to achieve TIR-specific representations by transferring the RGB model. Specifically, the target center location (TCL) distillation $\phi$ is used to identify the target from the distractors in the background for locating the position. Furthermore, to refine the bounding box of the results, the bounding box estimation (BBE) distillation $\theta$ is used to learn the regression coefficients to fit the target more accurately. 
Thus, the final distillation loss contains two parts: target center location loss and bounding box estimation loss, which can be written as: 
\vspace{-1mm}
\begin{equation}
\vspace{-1mm}
\setlength{\abovedisplayskip}{5pt}
\setlength{\belowdisplayskip}{5pt}
\label{eq:loss}
\mathcal L (\phi, \theta)=  \mathcal L_{TCL}(\phi) + \lambda \mathcal L_{BBE}(\theta),
\end{equation}
where $\lambda$ is a regularization coefficient to balance the two losses.

\vspace{-3mm}
\subsection{Cross-Modal Distillation}
\label{sec:mdt}
In this section, we introduce distillation modules between RGB and TIR modalities based on a large amount of paired RGB-TIR data.

\vspace{-1mm}
\minisection{Paired Patches Generation:}
The large amounts of annotated data are an essential factor for training network~\cite{imagenet} that has been demonstrated in other vision tasks. 
But for TIR tracking, lacking of large-scale training data limits the training of the network from scratch. 
Recent works~\cite{MLSSNet2020, MMNet2020} directly employ the pre-trained model in the RGB modality for the representations of TIR targets. 
The results show that this off-the-shelf representation is not optimal for the TIR target, as the appearance of the target in different modalities varies largely. The reason owes to the different imaging mechanisms.

Unlike the previous works~\cite{MLSSNet2020, MMNet2020}, 
in this work, we consider using a large amount of unlabeled paired data from RGB and TIR modalities in this study.
This avoids a lot of time and efforts to label them for the acquisition of large amounts of infrared data.
To better and more effectively utilize the paired RGB-TIR data for training, three methods: `center area', `random sampling', and `detection', are explored to generate the paired patches with same semantics as the input of the network.
Specifically,  the `center area' is to assign a square region in the center of the image as a fake target, thus obtaining the bounding box in this image.
The `random sampling' is to randomly sample several patches from the image and then they can be fed into the network for training. 

Additionally, we also employ the object detectors, such as YOLO~\cite{YoLOv4}, Faster R-CNN~\cite{FasterRCNN}, to detect the locations of the targets in the image.
After we obtain bounding boxes of `objects' by the above three methods, we can feed them for two applications in the training procedure. On one hand, the paired patches from dual-modalities are used as input of the network to execute forward inference. 
On the other hand, the corresponding coordinates of the patches are used to construct the distillation losses for back-propagation training.

\vspace{-1mm}
\minisection{Target Center Location Distillation:} 
The target center location (TCL) distillation is utilized to locate the target coarsely by predicting the score of heatmap in the image. 
This distillation pays more attention to robustness than accuracy during tracking.
In this distillation, we expect to learn a TIR filter $\phi$ derived from the TIR images while fitting with the RGB  representation with a Gaussian pseudo distribution $g$. The cross-modal distillation (CMD) loss of target center location can be written as follows:
\vspace{-1mm}
\begin{equation}
\vspace{-1mm}
\setlength{\abovedisplayskip}{3pt}
\setlength{\belowdisplayskip}{3pt}
\label{eq:loss1}
\mathcal L_{TCL}(\phi) = \sum_{i=1}^N||\phi(x_i^t) \otimes x_i^v - g||^2 +\mu||\phi||^2,
\end{equation}
where the filter is denoted as $\phi$,  $ \otimes $ is the convolution operation and $\mu$ is a  regularization parameter.  

This loss can be minimized to optimize the weights of a linear convolutional layer. 
It is helpful to identify the target from distractors in the background. 
During training, this branch is trained by a meta-learning way with the pre-trained RGB model using the above loss in Eq. ~\ref{eq:loss1}.
During the tracking inference,  the center position of the target is optimized by searching the maximum confidence score within a wide search region in the next frame. 

\vspace{-1mm}
\minisection{Bounding Box Estimation Distillation:} The BBE enables the tracker to be wrapped by the box for accuracy improvement of the tracking performance. 
Here, we denote the state of the bounding box as $s = (c_x/w, c_y/h, \log w, \log h) \in \mathbb R^4$.
The distillation loss of bounding box estimation can be denoted as:
\vspace{-1mm}
\begin{equation}
\vspace{-1mm}
\setlength{\abovedisplayskip}{3pt}
\setlength{\belowdisplayskip}{3pt}
\label{eq:loss2}
\mathcal L_{BBE}(\theta) = \sum_{i=1}^N||\psi(\theta(x_i^t)\odot\varphi(x_i^v)) - s||^2 +\nu||\theta||^2,
\end{equation}
where $\psi(\cdot)$ is the fully connection operation, $\odot$ is the pixel-wise multiplication, and $\varphi$ is pre-trained model in RGB modality.

We obtain the RGB representation by using the $\varphi$ to extract on the RGB features $x^v$, and regard it as the teacher vector to guide the student model  $\theta$ for learning TIR-specific representations under the pseudo label $s$.   
During tracking, the BBE  distillation is used to find the bounding box fitting the target ultimately by maximizing the intersection over union (IoU) scores. 

\vspace{-1mm}
\minisection{Unsupervised Training:}
In this work, we propose a representation transferring mechanism for two-branch trackers by cross-modal distillation learning.
Different from previous training approaches, the proposed training mechanism can work on a large amount of unlabeled RGB-TIR data with an unsupervised training manner. 
Then RGB and TIR CNN features extracted by the backbone network are utilized as the input of the Cross-Modal Distillation (CMD) for representation distillation. 
Specifically, RGB and TIR features extracted by the backbone are utilized as the input of the cross-modal Distillation (CMD) module. For the distillation mechanism, the RGB representation obtained by the pre-trained model plays the role of the teacher model, while the model for learning the TIR representation is the student model. 

As for the consistency of the semantics information between the paired RGB and TIR patches, the distillation training enables the TIR representation of the student model approachable to the RGB representation of the teacher model. Obviously, the training procedure mainly relies on the consistent information of the paired patches, thus the network can be trained in an unsupervised way. 
That is to say, our proposed method only needs the clean unlabeled paired dual-modalities images, avoiding the time-consuming, labor-intensive, and cumbersome manually labeled data.
Furthermore, it is easy to extend to an even larger scale of unlabeled training data.

\vspace{-2mm}
\section{Experiments}
\label{sec:exp}
In this section, we provide the experimental results of the proposed CMD method on LSOTB-TIR dataset~\cite{LSOTB} and PTB-TIR dataset~\cite{PTB-TIR} to verify its effectiveness. Besides, we compare the trackers equipped with our CMD with several state-of-the-art trackers.

\vspace{-1mm}
\subsection{Evaluation Datasets and Protocols}
\noindent \textbf{LSOTB-TIR dataset~\cite{LSOTB}}
is a large-scale high-diversity TIR tracking benchmark with a total of $1,400$ TIR sequences and more than $600$K frames. It is annotated with more than $730$K bounding boxes in total. The training dataset contains $1,280$ sequences with $47$ objects classes and over 650k bounding boxes. And it selects $120$ sequences, with $22$ object classes and more than $82$K frames, as the evaluation dataset. At present, it is larger and more diverse than other existing TIR datasets. We use the Precision, the Normalized Precision and the Success as the metrics for this evaluation dataset.

\noindent \textbf{PTB-TIR dataset~\cite{PTB-TIR}} is a TIR pedestrian tracking dataset for the TIR pedestrian tracker evaluation, which includes 60 thermal sequences with manual annotations. Each sequence has nine attribute labels for the attribute based evaluation to ensure the diversity of the dataset, and all of them come from different devices, scenes, and shooting times. The center location error (CLE) and overlap ratio (OR) are exploited as metrics. That is to say, the Precision Plot and Success Plot are used to rank trackers.

\noindent \textbf{Evaluation protocols.}
We use one-pass evaluation method~\cite{OTB} (OPE) that each tracker is only initialized in the first frame, and is not affected by the true position of the target during the entire tracking process. 
The performance of the algorithm is evaluated by precision, normalized precision and success rates. 
The center location error (CLE) refers to the Euclidean distance between the center of the predicted position and the center of the artificial mark. The Precision is the ratio of the number of video frames whose CLE is less than a given threshold to the total number of video frames. A threshold of $20$ pixels is usually set as the sorting criterion.
As the Precision is affected by the image's resolution and the size of the bounding box, we normalize the Precision over the size of the bounding box. Then, the area under the curve (AUC) of the Normalized Precision between $0$ and $0.5$ is used to evaluate the performance of the trackers.
The overlap rate (OR) is the ratio of the union and intersection of the predicted target area and the ground truth area. The Success is the ratio of the number of frames with an overlap rate greater than a set threshold ([$0,1$]) to the total number of frames. We usually use the area under the curve (AUC) as an indicator to measure the overall effectiveness of the tracking algorithm.

\newcommand{\tabincell}[2]{\begin{tabular}{@{}#1@{}}#2\end{tabular}} 
\begin{table*}[!htbp]
	\centering
	\begin{center}
		\vspace{-4.5mm}
		\caption{\small \textbf{Analysis of our cross-modal distillation on LSOTB-TIR~\cite{LSOTB} dataset.} We evaluate several variants of our proposed method based on DiMP~\cite{DIMP}. The best results are highlighted in bold font.}
		\vspace{-4mm}
		\label{DiMP}
	\end{center}
	{
		\resizebox{.9\textwidth}{!}{
			\begin{tabular}{l|l|cc|ccc}
				\hline
				\tabincell{l}{Modules to \\ be Updated} &  Training Settings & Reference Branch & Test Branch & Success($\uparrow$) & Precision($\uparrow$) & Normalized Precision($\uparrow$) \\
				\hline
				None(Baseline) & A1. Same as DiMP~\cite{DIMP}&  RGB &  RGB & 66.2 (0.0) & 78.7 (0.0) & 70.7 (0.0)\\
				\hline
				\multirow{3}*{BBE} 
				& B1. Learning on reference branch & TIR(ft) &  RGB & 67.4 (1.2) & 80.4 (1.7) & 72.1 (1.4)\\
				& B2. Learning on test branch & RGB & TIR(ft) & 67.2 (1.0) & 80.1 (1.4) & 71.9 (1.2)\\
				& \tabincell{l}{B3. Combination of B1 and B2} & TIR & TIR & 67.8 (1.6) & 80.8 (2.1) & 72.5 (1.8)\\
				\hline
				TCL & C1. Learning on reference branch &  TIR(ft) & RGB & 67.7 (1.5) & 80.7 (2.0) & 72.5 (1.8) \\
				\hline
				\multirow{5}*{BBE and TCL} 
				& D1. Joint learning on test branch & RGB & TIR(ft) & 67.1 (0.9) & 79.7 (1.0) & 71.8 (1.1)\\
				& D2. Joint learning on reference branch & TIR(ft) & RGB & 67.6 (1.4) & 80.6 (1.9) & 72.3 (1.6)\\
				& D3. Joint learning & RGB-TIR(ft) & TIR-RGB(ft) & \textbf{68.0 (1.8)} & \textbf{80.8 (2.1)} & \textbf{72.7 (2.0)}\\
				& D4. Joint learning with random sampling & RGB-TIR(ft) & TIR-RGB(ft) & 67.3 (1.1) & 80.2 (1.5)& 72.1 (1.4)\\ 
				& D5. Joint learning with a detector & RGB-TIR(ft) & TIR-RGB(ft) &  67.5 (1.3) &  80.4 (1.7) & 72.3 (1.6)\\
				\hline
		\end{tabular}}
	\vspace{-3mm}
	}
\end{table*}

\vspace{-2mm}
\subsection{Implementation Details}
For the baseline tracker DiMP~\cite{DIMP}, we use the default settings, referring to details in the paper~\cite{DIMP}. 
Here, we update Target Center Location (TCL) and Bounding Box Estimation (BBE) modules to carry out the cross-modal distillation during training.
TIR images lack details and texture information compared with RGB images, while the TIR target's appearance changes stably during tracking. 
Thus, we need to update the tracking model slightly and carefully to adapt to the TIR characteristics. 
In practice, we reduce the learning rates used to update BBE and TCL, both to 1e-6 as the rows \textbf{B} \& \textbf{C} in Table~\ref{DiMP}.
Considering the difference of convergence rates between BBE and TCL, we set the learning rates of BBE and TCL as $1e-7$ and $2e-8$ respectively, during the joint distillation as the row \textbf{D} in Table~\ref{DiMP}. 
For fair comparison, the model is trained for $50$ epochs with mini-batch size of $5$, and the learning rate is decreased to multiples of $0.5$ at every $15$ epochs in all experiments. 
The number of samples in every epoch is $26,000$.

From the training datasets, we sample the paired RGB and TIR patches with the same semantics.
RGB patches are fed into one branch of the network and the corresponding TIR patches are input for the other branch. 
Specifically, we input $5$ RGB images and $5$ TIR images for each branch.
In addition, in order to enhance the mutual information between the modalities during the cross-modal distillation, we connect the paired RGB and TIR patches along the horizontal direction in spatial domain as the input of the network. 
The connected patch `RGB-TIR' means that the RGB patch is in the left side of the TIR patch, while vice versa is called `TIR-RGB'.
Then, they are fed to the two branches for distilling the representation.
In the next part, we describe our three approaches to generate the paired patches which are aligned strictly with the same semantic information from above paired RGB and TIR images.

\noindent \textbf{Paired Patches Generation.}
We use the training data from the work~\cite{ECO-stir} which takes advantage of large-scale paired RGB-TIR data by collecting from several datasets, e.g. KAIST dataset~\cite{KAIST}. The whole training data consists of $126,666$ paired RGB and TIR images which are unlabeled for tracking.

We propose three approaches to implement the generation of paired patches from the above training data. 
The first is to assign central regions of the paired images, which is regarded as a fake object, to be the input paired patches. We call this method as `center area'. 
The second is to randomly crop some regions of approximately 1/6 size of the image, and then resize these regions to $100 \times 100$ pixels as the patches. We call this method as `random sampling'.
In order to obtain more accurate location of objects on RGB-TIR datasets, we use an object detector as the third method called `detection'. In this way, the size of the paired patches is dependent on the detection results. Specifically, as the detector is trained in RGB modality, it is suitable for the RGB detection. So, we use it to detect objects of RGB images.

\vspace{-2.5mm}
\subsection{Analysis of Distillation Mechanisms}
For the training settings in DiMP~\cite{DIMP}, the inputs of reference and test branches are temporal patches from the same sequence. That is to say, they aim to make full use of the target's change in the single modality based on the continuity of time. Different from that, our goal is to learn cross-modal knowledge by transferring the high-level semantics of the same object in the RGB modality to the TIR modality. After the distillation, the input of the tracking process is only from the TIR modality.

Table~\ref{DiMP} shows our analysis of the effectiveness of cross-modal distillation to Target Center Location (TCL) and Bounding Box Estimation (BBE) over the baseline tracker DiMP~\cite{DIMP}. 
The tracking results of the baseline tracker are presented as \textbf{A1}.

All the experiments mainly focus on updating the parameters in the branch which is input with TIR patches. 
We train two modules for the DiMP~\cite{DIMP}, including the BBE and the TCL. The results are reported in terms of Success, Precision, and Normalized Precision. We explicitly show the patches with the corresponding modality attribute for the reference branch and the test branch of the tracker. Here, `RGB' and `TIR' mean that the patch is from a single modality.

Then, they are mixed to do cross-correlation in the Siamese architecture. For example, for training the reference branch, we keep the upper half and the lower half of a mini-batch as patches from RGB modality and TIR modality respectively, namely `RGB-TIR' in the table. Thus the corresponding places in mini-batch for test branch are input with patches from TIR modality and RGB modality respectively, namely `TIR-RGB' in the table. `ft' means parameters of the branch are to be fine-tuned in the network.

\noindent \textbf{Bounding Box Estimation (BBE).}
For the {BBE}, we consider three cases, namely `B1. Learning on reference branch', `B2. Learning on test branch' and `B3. Combination of B1 and B2'.

\begin{figure*}
	\subfigure{\includegraphics[width=0.30\textwidth]{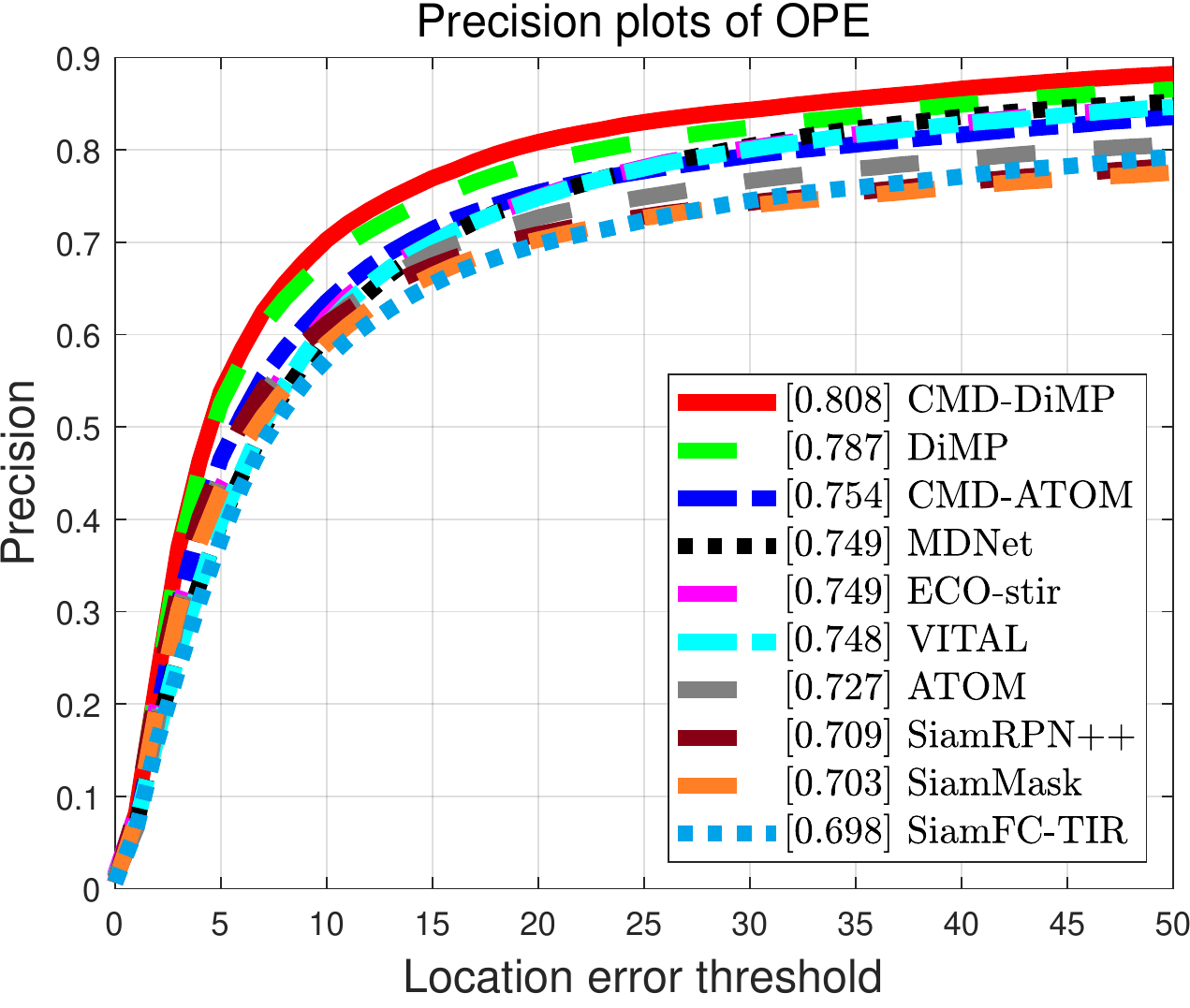}}
	\subfigure{\includegraphics[width=0.30\textwidth]{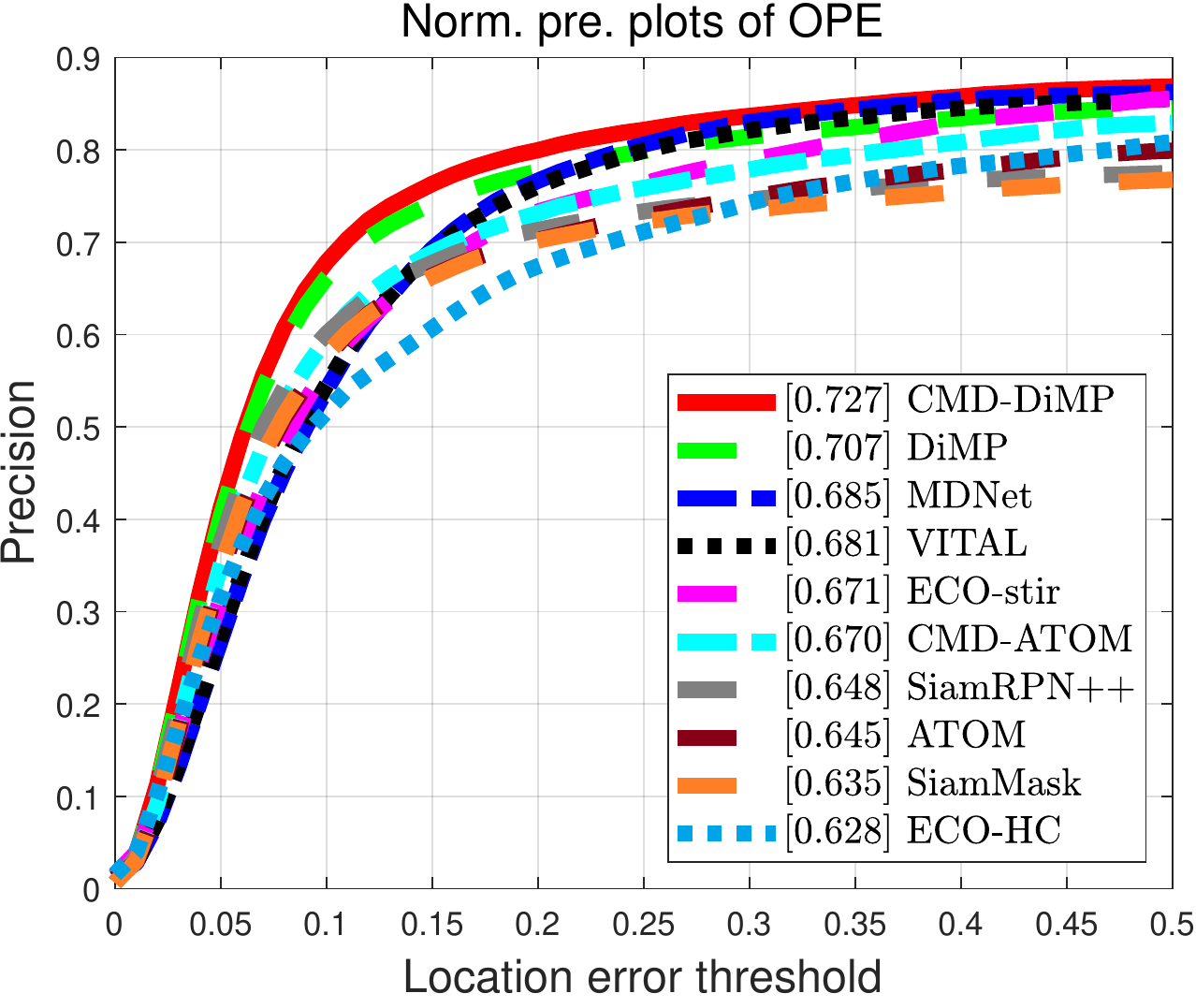}}
	\subfigure{\includegraphics[width=0.30\textwidth]{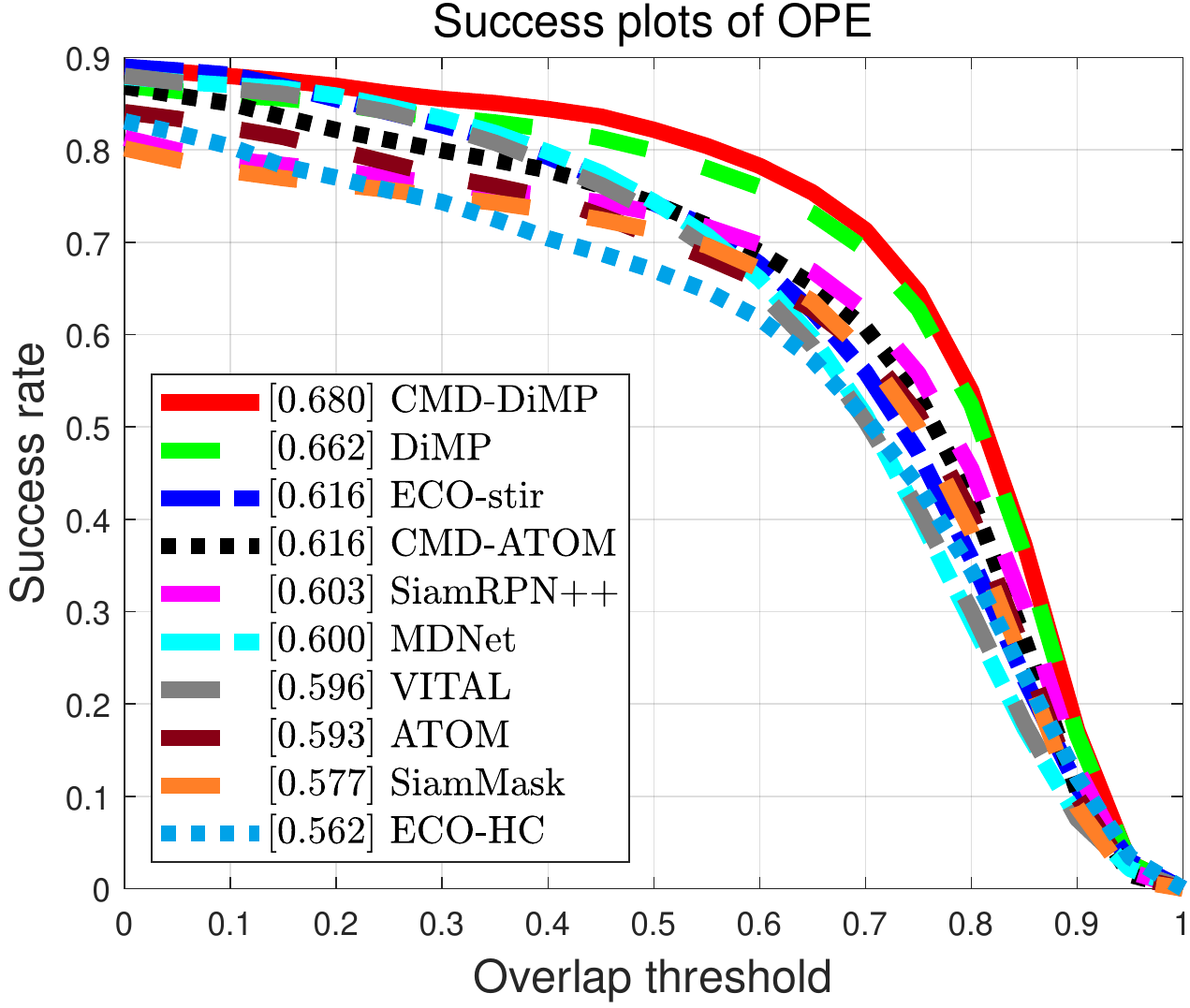}}
	\centering
	\vspace{-5mm}
	\caption{State-of-the-art comparison on LSOTB-TIR dataset~\cite{LSOTB}. We compare our method with top-10 trackers in terms of Precision, Normalized Precision, and Success. Our proposed approach achieves the best performances on three metrics.}
	\label{lsotb}
	\vspace{-4mm}
\end{figure*}

\vspace{-1mm}
\begin{itemize}
	\item \textbf{B1} We input cross-modal patches to the Siamese branches of BBE. Specifically, we input the TIR patches to the reference branch and input the RGB patches to the test branch in the BBE, separately. We only update the parameters in the model of the TIR branch, namely the reference branch.
	Thus, we use the well-trained model in the tracking process. Compared with the original model, only the reference branch is updated. We input the TIR patches from the testing dataset~\cite{LSOTB} to the well-trained model.
	This method improves $1.2\%$ in Success, $1.7\%$ in Precision, and $1.4\%$ in Normalized Precision. 
	These results indicate that the reference branch needs to be trained to fit the TIR modality to obtain a better representation.
	
	\item \textbf{B2} We flip the input of patches from the two modalities. Specifically, the TIR patches are input to the test branch and the RGB patches are input to the reference branch.
	Therefore, the parameters of the test branch in the model are updated by the distillation 
	Similarly, as the \textbf{B1} method, we only update the parameters of the TIR branch. Both of the two methods improve the performance of the baseline tracker. Transfer between RGB and TIR modalities for BBE is useful and any branch of BBE has its own function. Besides, the reference branch trained in \textbf{B1} plays a litter better than the test branch trained in \textbf{B2} with $0.2\%$ in terms of Success, $0.3\%$ in terms of Precision, and $0.2\%$ in terms of Normalized Precision. 
	
	\item \textbf{B3} We combine the information from the two modalities. In practice, we borrow the parameters of the reference branch trained in \textbf{B1} and also the parameters of the test branch trained in the \textbf{B2}. Then we recombine them together as the complete BBE. 
	We achieve the best result and our tracker outperforms baseline tracker $1.6\%$, $2.1\%$ and $1.8\%$ in terms of Success, Precision, and Normalized Precision respectively.
	\vspace{-1mm}
\end{itemize}

The results in case (\textbf{B1}) achieves better results than that in case (\textbf{B2}).
We think the reason is that the reference branch contains more operations and can transfer more representation knowledge from the RGB model than the test branch.
When the reference and test branches (\textbf{B3}) are combined,  we can achieve the best performance due to the iterative cross-modal distillation operation that can optimize the learned representation of TIR modality.

\noindent \textbf{Target Center Location (TCL).}
We analyze the cross-modal distillation on TCL as `C1. Learning on reference branch' in Tab.~\ref{DiMP}. As the model parameters of TCL mostly gather in the reference branch, we only update the parameters in the reference branch. Therefore, the input of the reference branch is the data from the TIR modality and that of the test branch is the data from the RGB modality. From Tab.~\ref{DiMP}, we can see that the tracker equipped with our cross-modal distillation mechanism performs better than the baseline tracker. With the setting \textbf{C1}, we improve the results of the tracker by $1.5\%$ in Success, $2.0\%$ in Precision, and $1.8\%$ in Normalized Precision, showing that we successfully transfer the RGB representation in the test branch to the reference branch in TCL.

By using cross-modal distillation on TCL, we obtain the tracking results which exceed the performances of the tracker equipped with the \textbf{B1} and \textbf{B2}. We attribute it to the reason that TCL is critical for the location of the target, while BBE is to further refine the target scale based on the location of TCL. All these analyses furthermore demonstrate that our cross-modal distillation can learn more effective TIR representations for TIR images.

\noindent \textbf{Joint Learning of BBE and TCL.}
Inspired by the combination of the learned {reference branch and test branch in BBE} as \textbf{B3}, we attempt to jointly train BBE and TCL. We consider five cases, i.e. `D1. Joint learning on test branch', `D2. Joint learning on reference branch', `D3. Joint learning', `D4. Joint learning with random sampling' and `D5. Joint learning with a detector', aiming to obtain optimized parameters for each of branches on both TCL and BBE.

We use parameters of the combination models on \textbf{B3} and \textbf{C1} for the initialization of our model for jointly training. 
At the beginning, we try to train one branch with TIR input by transferring from the other branch with RGB input for both TCL and BBE simultaneously, i.e. case (\textbf{D1}) and case (\textbf{D2}). We improve the performances of the baseline tracker in both two cases.

In order to better train the model jointly and avoid the simple combination of two branches in TCL and BBE, we mix patches from both RGB and TIR modalities and then input them to the reference branch and also the test branch as in \textbf{D3}. The results of this method achieve 68.0\% in Success, 80.8\% in Precision, and 72.7\% in Normalized Precision, which are the best results at present.
All above cases, i.e. \textbf{B1}, \textbf{B2}, \textbf{B3}, \textbf{C1}, \textbf{D1}, \textbf{D2} and \textbf{D3}, use the central regions in the RGB and TIR images for generating the patches.

Besides, we propose randomly assigning different regions in an image to be the input patches as \textbf{D4}. Then we can obtain the paired patches under the alignment restriction of the paired images. For this case, we improve the performance by $1.1\%$ on Success, $1.5\%$ on Precision, and $1.4\%$ on Normalized Precision.

Additionally, we explore to use a detector to generate the input patches for distillation training as \textbf{D5}. But the classes of generated patches are restricted to the pre-trained detector model, which normally detects objects with specific classes such as `person', `car' etc. Therefore, the results obtained by setting with a detector, namely case (\textbf{D5}), are not good enough, compared with results of case (\textbf{D3}). As CMD with a detector is a bit better than that with randomly sampling as in Tab.~\ref{DiMP}, we attribute it to that `random sampling' could not contain any complete object contents of the image.

For now, we can summarize that input patches generated by a detector and random sampling are both restricted to the varieties of the object classes of training data during cross-modal distillation.

\begin{table*}[htbp]
	\centering
	\begin{center}
		\vspace{-4mm}
		\caption{\small \textbf{Attributes analysis of cross-modal distillation on LSOTB-TIR dataset~\cite{LSOTB}.} We evaluate our trackers with several RGB and TIR trackers on Precision, Normalized Precision, and Success (P$/$NP$/$S $\%$). The best results are highlighted in bold font.}
		\vspace{-4mm}
		\label{Attribute}
	\end{center}
	{
		\resizebox{1\textwidth}{!}{
			\begin{tabular}{l|l||c|c|c|c|c||c}
				\hline
				\tabincell{l}{Attributes Type} & Attributes Name & SiamFC-TIR~\cite{SiamFCtir} & SiamRPN++~\cite{SiamRPNjiajia} & VITAL~\cite{VITAL} & ECO-stir~\cite{ECO-stir} & DiMP~\cite{DIMP} & CMD-DiMP \\
				\hline
				\multirow{12}*{Challenge} 
				& Deformation & 71.1/61.9/56.3 & 66.8/60.2/56.6 & 74.8/64.7/56.9 & 76.3/65.9/60.1 & 76.3/67.0/63.3 & \textbf{80.5/70.7/66.8}\\
				& Occlusion & 65.1/59.2/53.9 & 63.9/58.6/54.6 & 72.7/65.9/59.4 & 71.1/64.9/59.4 & 73.7/65.7/62.2 &  \textbf{76.2/67.8/64.0}\\
				& Distractor & 63.6/56.8/51.0 & 64.8/58.1/54.6 & 71.4/64.9/57.7 & 74.7/67.2/61.6 & 75.0/66.4/63.0 &  \textbf{76.1/67.3/63.7}\\
				& Background clutter & 69.2/62.3/54.6 & 70.4/65.1/60.6 & 73.0/67.2/58.4 & 73.9/67.2/61.3 & 78.9/70.9/66.4 &  \textbf{79.8/71.6/67.0} \\
				& Out of view & 73.0/65.9/58.6 & 75.0/70.7/63.9 & 70.6/68.5/59.7 & 76.7/72.1/66.8 & 77.4/74.0/67.4 &  \textbf{79.7/76.1/69.6}\\
				& Scale variation & 74.7/68.0/58.4 & 81.3/74.2/68.2 & 79.7/74.5/62.1 & 80.5/74.2/68.0 & 89.2/81.3/75.1 &  \textbf{91.0/83.0/76.7}\\
				& Fast motion & 72.4/68.8/59.7 & 74.8/70.3/64.4 & 72.2/67.8/58.2 & 73.8/68.0/63.6 & 82.7/77.1/71.1 &  \textbf{86.1/80.1/74.1}\\
				& Motion blur & 74.0/64.7/57.6 & 70.7/64.8/58.8 & 76.8/69.4/59.3 & 76.7/66.8/61.3 & 80.3/73.0/67.1 &  \textbf{85.6/77.7/71.5}\\
				& Thremal crossover & 68.0/55.7/51.7 & 60.0/50.2/48.4 & \textbf{78.0/64.3/58.1} & 73.1/58.6/54.5 & 61.9/50.6/50.2 &  64.7/54.2/52.9\\
				& Intensity variation & 85.0/77.7/71.6 & 83.6/76.7/74.4 & 74.1/72.5/61.8 & 76.9/76.1/70.9 & 91.4/87.6/82.7 &\textbf{91.5/87.7/82.8}\\
				& Low resolution & 91.4/74.2/65.2 & 81.2/66.3/62.0 & 91.1/73.6/60.9 & \textbf{94.1/69.7/64.0} & 78.1/64.6/60.3 & 83.1/68.4/64.0 \\
				& Aspect ratio variation & 69.5/59.8/48.9 & 70.6/65.0/59.4 & 72.9/63.7/54.2 & 72.4/58.1/54.9 & 78.9/72.7/\textbf{67.2} & \textbf{80.4/70.3}/65.0 \\
				\hline
				\multirow{4}*{Scenario} 
				& Vehicle-mounted & 74.5/70.1/59.4 & 86.0/79.2/72.6 & 83.7/81.5/72.1 & 84.4/81.1/76.0 & 91.9/84.2/78.8 & \textbf{96.6/88.3/82.4}\\
				& Drone-mounted & 68.2/60.2/53.7 & 64.9/58.4/55.0 & 69.5/64.1/53.8 & 69.5/61.5/55.7 & \textbf{75.5/69.7/64.1} & 73.0/67.4/62.0\\
				& Surveillance & 63.9/58.0/53.5 & 67.0/61.7/57.7 & 70.0/63.5/57.5 & 69.4/64.1/59.1 & 74.5/\textbf{65.4/62.9} & \textbf{74.8}/65.3/62.8\\
				& Hand-held & 74.8/64.6/56.3 & 70.7/64.7/59.8 & 78.8/68.3/58.8 & 79.4/66.4/60.3 & 78.0/69.9/64.4 & \textbf{84.1/75.8/69.9}\\
				\hline
				\tabincell{l} All & All & 69.8/62.4/55.3 & 70.9/64.8/60.3 & 74.8/68.1/59.6 & 74.9/67.1/61.6 &  78.7/70.7/66.2 & \textbf{80.8/72.7/68.0} \\
				\hline
		\end{tabular}}
		\vspace{-4mm}
	}
\end{table*}

\begin{figure}[htbp]
\vspace{-3mm}
	\begin{minipage}[t]{0.23\textwidth}
		\centering
		\includegraphics[width=4.0cm]{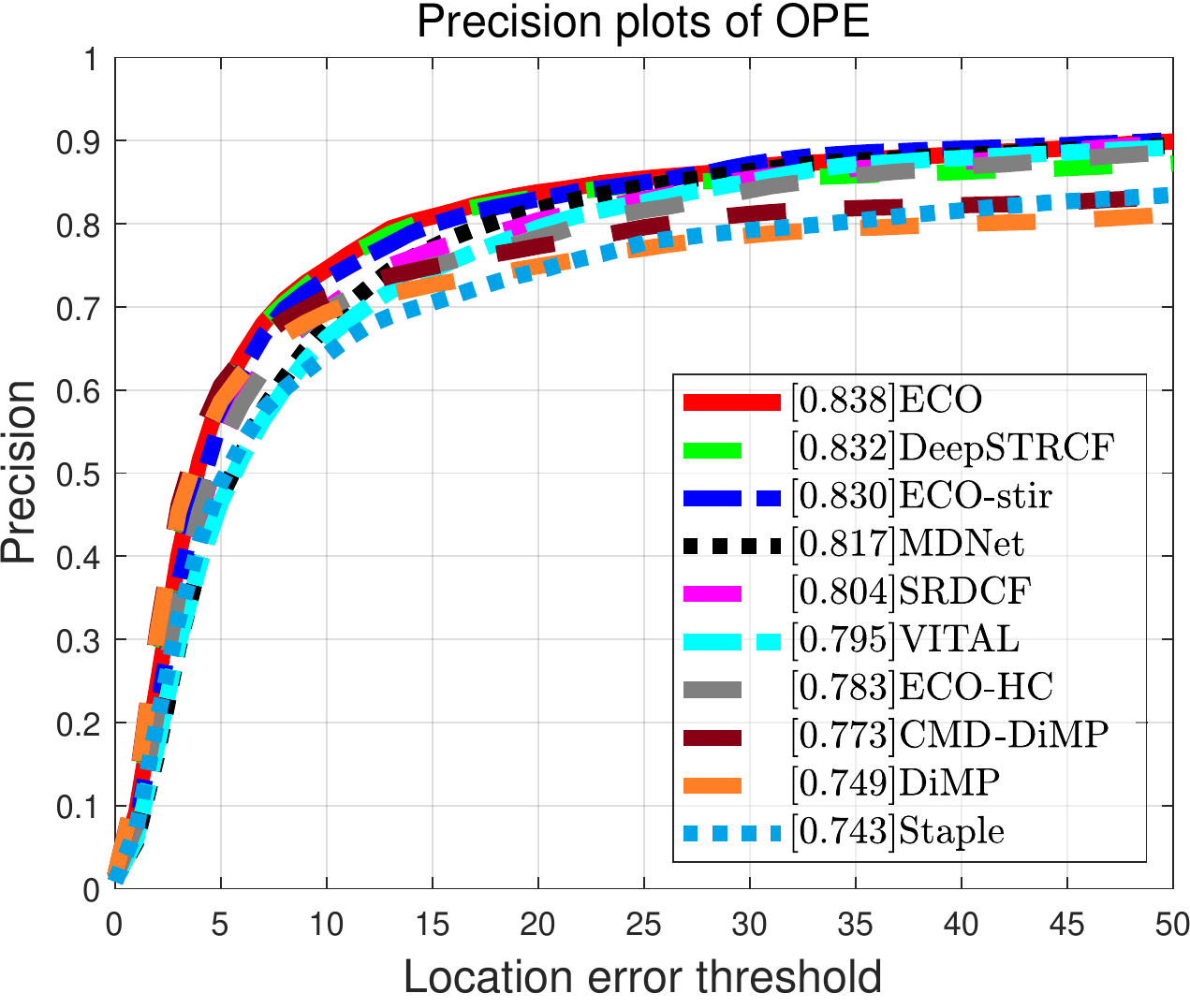}
	\end{minipage}
	\begin{minipage}[t]{0.23\textwidth}
		\centering
		\includegraphics[width=4.0cm]{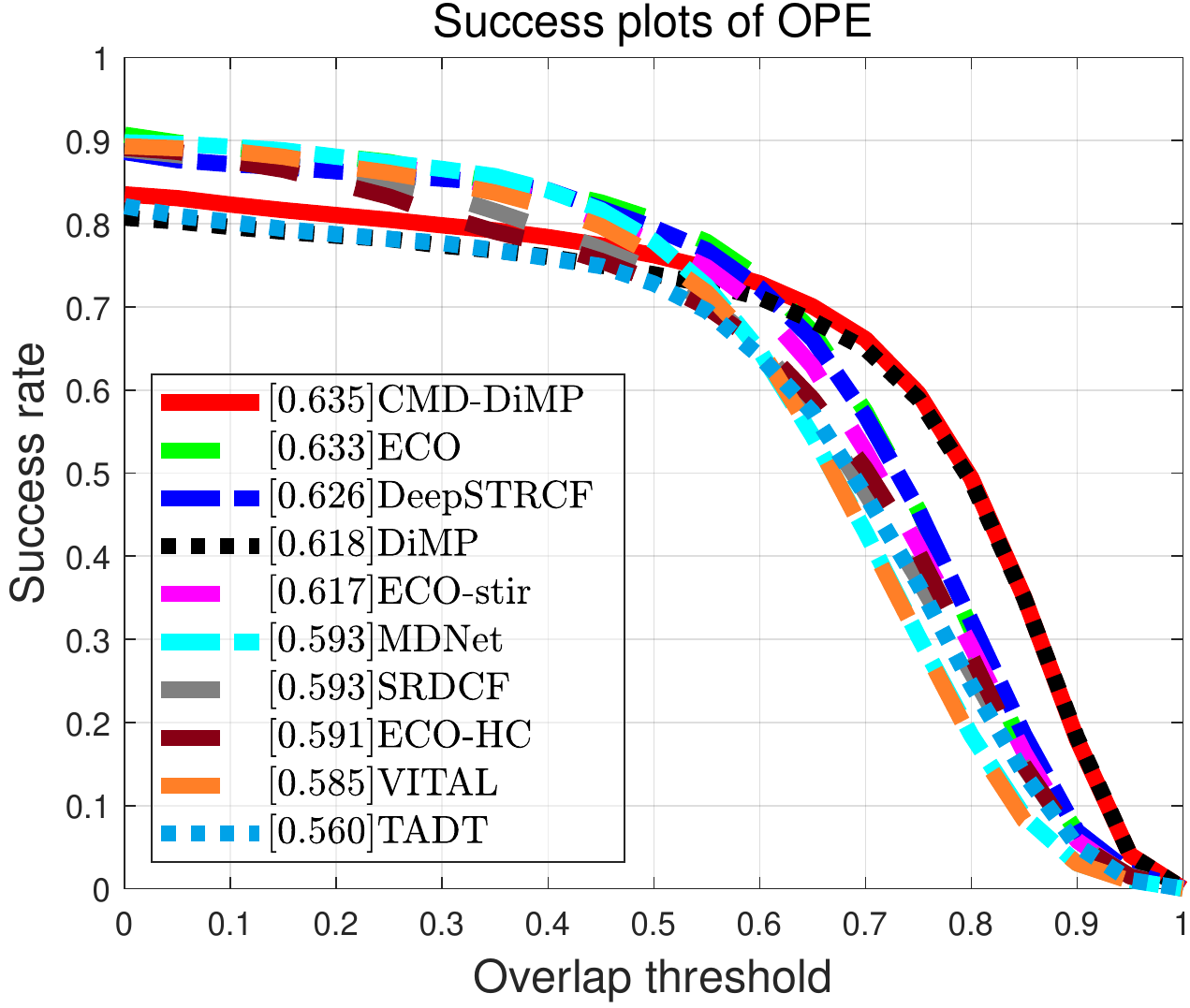}
	\end{minipage}
	\vspace{-3mm}
	\caption{Precision plot and success plot by comparing our tracker with the top-10 trackers on PTB-TIR dataset~\cite{PTB-TIR}.}
	\label{ptbtir}
	\vspace{-4mm}
\end{figure}

\vspace{-2mm}
\subsection{LSOTB-TIR Dataset}
\label{4_exp_lsotb}
\vspace{0mm}
In this part, we compare our tracker (CMD-DiMP) against high-quality trackers on the LSOTB-TIR~\cite{LSOTB} dataset in terms of Success, Precision and Normalized Precision in Fig.~\ref{lsotb}, including TIR tracker,~\emph{e.g.} ECO-stir~\cite{ECO-stir}, and RGB trackers:~\emph{e.g.} MDNet~\cite{MDNet}, VITAL~\cite{VITAL}, SiamRPN++~\cite{SiamRPNjiajia}, SiamMask~\cite{SiamMask}, SiamFC-TIR~\cite{SiamFCtir}, ECO-HC~\cite{ECO} and so on. By using our cross-modal training, we improve baseline tracker~\cite{DIMP} with absolute gains of $1.8\%$, $2.1\%$ and $2.0\%$ in terms of Success, Precision and Normalized Precision respectively. These obvious improvements prove that our cross-modal training can extract more useful and consistent information between RGB and TIR modalities.

Besides, we extend our cross-modal distillation training mechanism to the variant tracker ATOM~\cite{ATOM} to prove our method's generality. For the original ATOM~\cite{ATOM} tracker, only the IoUNet component is updated during offline training. Therefore, here we only employ the cross-modal distillation on the Bounding Box Estimation (BBE).
We evaluate our tracker (CMD-ATOM) on the LSOTB-TIR dataset~\cite{LSOTB}. And we improve the baseline tracker ATOM~\cite{ATOM} with absolute gains of $2.3\%$, $2.7\%$, and $2.5\%$ in Success, Precision, and Normalized Precision respectively, shown in Fig.~\ref{lsotb}.

\vspace{-2mm}
\subsection{PTB-TIR Dataset}
\vspace{0mm}
We evaluate our tracker (CMD-DiMP) on the PTB-TIR dataset~\cite{PTB-TIR} using the two evaluation metrics, i.e. Precision and Success. We compare our tracker with some trackers on PTB-TIR dataset~\cite{PTB-TIR}, including TIR tracker ECO-stir~\cite{ECO-stir}, and RGB trackers, such as ECO~\cite{ECO}, DeepSTRCF~\cite{DeepSTRCF}, MDNet~\cite{MDNet}, SRDCF~\cite{SRDCF},  VITAL~\cite{VITAL}, Staple~\cite{Staple}, TADT~\cite{TADT}, MCCT~\cite{MCCT}, ECO-HC~\cite{ECO} and so on.

As shown in Fig.~\ref{ptbtir}, our tracker gets $0.773$ in terms of Precision and $0.635$ in terms of Success, which improves $2.4\%$ and $1.7\%$ for the baseline tracker separately. We win first place in the Success plot and get a lower score in the Precision plot. We observe that the predicted bounding boxes will be expanded to a larger area for containing more similar areas, such as the distractors or background clutter, which reduces the Precision drastically.
More importantly, as the BBE pays more attention to the object's scale changes, there is an obvious improvement when the overlap threshold is bigger than $0.6$ in the Success plot of OPE.

\vspace{-1.5mm}
\subsection{Attribute Analysis on LSOTB-TIR Dataset}
\vspace{0mm}
The LSOTB-TIR~\cite{LSOTB} dataset provides $16$ attributes to be evaluated, including $12$ challenges and $4$ scenarios. We compare our tracker (CMD-DiMP) with TIR trackers, ECO-stir~\cite{ECO-stir} and SiamFC-TIR~\cite{SiamFCtir}, and RGB trackers, VITAL~\cite{VITAL}, SiamRPN++~\cite{SiamRPNjiajia} and DiMP~\cite{DIMP} in terms of Precision, Normalized Precision and Success (P$/$NP$/$S $\%$). Table~\ref{Attribute} shows that our tracker outperforms the above five trackers for most of the challenges except thermal crossover, low resolution {and drone-mounted.}

We improve the performance by about $1\%$-$6\%$ in most of attributes compared with DiMP~\cite{DIMP}.
{Compared with other four trackers, we obtain significant progress in most challenges.}
For attributes like deformation, background clutter, and aspect ratio variation, we get over $6\%$ absolute gains in terms of Success score. 
Even more, we make progress of $8\%$ in terms of Success score for attributes of scale variation and intensity variation.
And for challenges such as fast motion and motion blur, our tracker achieves $10\%$ improvement on Success score.
Besides, for vehicle-mounted and drone-mounted, the improvement on Success score made by our tracker is up to $6\%$ and that for hand-held is near to $10\%$.

\vspace{-1mm}
\section{Conclusions}
\label{sec:con}
In this paper, we propose to learn TIR-specific representations by distilling the pre-trained model of RGB modality to TIR modality on unlabeled RGB-TIR datasets, called cross-modal distillation (CMD).
During distillation, representations from the RGB modality model are used as supervised signals to guide the learning of  TIR-specific representations for the TIR modality. 
We take advantage of two branches of the network to deal with data from different modalities for transferring cross-modal knowledge.
Moreover,  we explore the prior semantic consistency implied in the training data structure itself to automatically generate paired RGB-TIR patches for training. 
Unlike the normal RGB tracking datasets, the training data for our method can be without any annotation about the targets.
Our cross-modal distillation mechanism reduces the dependency of training networks on labeled data and increases the extendibility with a larger amount of unlabeled data in the future.
Through this mechanism, we leverage the pre-trained RGB models which are derived from large-scale annotated RGB data, to train models specific for representing targets in TIR modality in an unsupervised way.  
Extensive experimental results on two standard TIR datasets demonstrate that our proposed CMD effectively improves the performance of the baseline tracker for TIR tracking.

\begin{acks}
This work is supported in part by National Natural Science Foundation of China (61773397,62006245, 61971352,61862043,U19B2037).
\end{acks}

\clearpage

\bibliographystyle{ACM-Reference-Format}
\balance
\bibliography{sample-base}

\end{document}